\documentclass[journal]{IEEEtran}

\usepackage{ifpdf}

\usepackage{cite}

\ifCLASSINFOpdf
  \usepackage[pdftex]{graphicx}
\else
  \usepackage[dvips]{graphicx}
\fi

\usepackage{amsmath}
\usepackage{algorithmic}

\usepackage{array}

\ifCLASSOPTIONcompsoc
  \usepackage[caption=false,font=normalsize,labelfont=sf,textfont=sf]{subfig}
\else
  \usepackage[caption=false,font=footnotesize]{subfig}
\fi

\usepackage{fixltx2e}

\usepackage{stfloats}
\ifCLASSOPTIONcaptionsoff
  \usepackage[nomarkers]{endfloat}
 \let\MYoriglatexcaption\caption
 \renewcommand{\caption}[2][\relax]{\MYoriglatexcaption[#2]{#2}}
\fi
\let\MYorigsubfloat\subfloat
\renewcommand{\subfloat}[2][\relax]{\MYorigsubfloat[]{#2}}
\usepackage{url}

\usepackage{hyperref}
\hypersetup{colorlinks=true, linkcolor=blue, filecolor=cyan, urlcolor=magenta}
\usepackage{booktabs}
\usepackage{multirow}
\usepackage{amsfonts}
\usepackage{amssymb}
\usepackage[linesnumbered, lined, ruled, commentsnumbered]{algorithm2e}
\usepackage{threeparttable}
\usepackage[switch]{lineno}

\begin{document}
\title{Driving Behavior Modeling using Naturalistic Human Driving Data with Inverse Reinforcement Learning}

\author{Zhiyu Huang,~\IEEEmembership{Student Member,~IEEE},
        Jingda Wu,~\IEEEmembership{Student Member,~IEEE}, 
        Chen Lv,~\IEEEmembership{Senior Member,~IEEE}
\thanks{Z. Huang, J. Wu and C. Lv are with the School of Mechanical and Aerospace Engineering, Nanyang Technological University, Singapore, 639798. (e-mail: zhiyu001@e.ntu.edu.sg, jingda001@e.ntu.edu.sg, lyuchen@ntu.edu.sg)}%

}
\maketitle

\begin{abstract}
Driving behavior modeling is of great importance for designing safe, smart, and personalized autonomous driving systems. In this paper, an internal reward function-based driving model that emulates the human's decision-making mechanism is utilized. To infer the reward function parameters from naturalistic human driving data, we propose a structural assumption about human driving behavior that focuses on discrete latent driving intentions. It converts the continuous behavior modeling problem to a discrete setting and thus makes maximum entropy inverse reinforcement learning (IRL) tractable to learn reward functions. Specifically, a polynomial trajectory sampler is adopted to generate candidate trajectories considering high-level intentions and approximate the partition function in the maximum entropy IRL framework. An environment model considering interactive behaviors among the ego and surrounding vehicles is built to better estimate the generated trajectories. The proposed method is applied to learn personalized reward functions for individual human drivers from the NGSIM highway driving dataset. The qualitative results demonstrate that the learned reward functions are able to explicitly express the preferences of different drivers and interpret their decisions. The quantitative results reveal that the learned reward functions are robust, which is manifested by only a marginal decline in proximity to the human driving trajectories when applying the reward function in the testing conditions. For the testing performance, the personalized modeling method outperforms the general modeling approach, significantly reducing the modeling errors in human likeness (a custom metric to gauge accuracy), and these two methods deliver better results compared to other baseline methods. Moreover, it is found that predicting the response actions of surrounding vehicles and incorporating their potential decelerations caused by the ego vehicle are critical in estimating the generated trajectories, and the accuracy of personalized planning using the learned reward functions relies on the accuracy of the forecasting model. Code is available at: \href{https://github.com/MCZhi/Driving-IRL-NGSIM}{https://github.com/MCZhi/Driving-IRL-NGSIM}. 
\end{abstract}


%
\IEEEpeerreviewmaketitle

\section{Introduction}
\label{sec1}
%
%
%
%
\IEEEPARstart{H}{uman-like} driving is an essential objective for autonomous vehicles (AVs) targeting widespread deployment in the real world. It is conceivable that AVs and human drivers share the road in the near future, which requires AVs to act like humans, thus being predictable and interpretable to other human drivers, in order to operate safely among humans. However, current AVs fail to show such characteristics, which would lead to conservative and unnatural decisions that may confuse and even endanger other human drivers \cite{li2018humanlike}. This is due to their inability to interact with other human traffic participants, or more specifically, to reason about other agents' possible behaviors and make proactive decisions accordingly. This problem motivates us to study and understand human driving behaviors, which is crucial to enable a safe and efficient autonomous driving system. Moreover, personalization should be another integral facet of human-like driving, which means the AV should make decisions according to the user's personal preferences \cite{huang2021personalized}. This motivates us to research individual driving behavior and thereby express human driving styles explicitly and individually.

Current development in artificial intelligence \cite{ma2020artificial} has provided us powerful tools in learning human driving behaviors for AV decision making, among which imitation learning is a state-of-the-art method to learn to make decisions by imitating human demonstration actions \cite{kebria2019deep, huang2020multi}. This paper adopts a similar technique but instead of learning decision-making policy directly, we learn the internal reward function used for decision-making. The prior assumption is that rational drivers choose actions that optimize their internal reward functions, and we choose the internal reward function approach to model driving behavior for the following reasons. First of all, this approach formulates the motivation of agents choosing actions and is believed to better reflect the human's internal decision-making scheme. Secondly, the form of a reward function (mostly linear) is highly succinct and interpretable because of its explicit physical meanings \cite{sun2019interpretable}, so that the weights of the reward function can be adjusted to explicitly reflect the preferences of different human drivers. To obtain the parameters of the reward function that best explains human behavior, inverse reinforcement learning (IRL) has emerged as a major approach. Specifically, maximum entropy IRL \cite{ziebart2008maximum} has received much attention in driving behavior modeling, thanks to its capability in addressing the stochasticity of driving behavior and the ambiguity that multiple reward functions can explain the human's behavior. However, the original setting of maximum entropy IRL is limited to discrete and small-scale problems as it relies on value iteration to evaluate the reward and state visitation frequency to calculate feature expectations, which are intractable to solve in high dimensional problems with large and continuous state space.

For our driving behavior modeling task, we should notice that humans' actions are actually governed by their latent states on high-level tactical intentions. Thus, we propose a structural assumption of human driving behavior that focuses on the discrete latent states instead of continuous states and actions, which makes it tractable to utilize the maximum entropy IRL framework to recover the underlying reward function. Particularly, we assume a human driver makes long-term decisions rather than instant actions according to three sequential processes, namely generation, evaluation, and selection. To put it simply, a human driver should first generate multiple candidate trajectories in mind, anticipate their outcomes, evaluate the rewards, and finally select one to follow. This structured assumption can make it tractable to calculate the partition function in the maximum entropy IRL framework and also significantly reduce the computation complexity. We validate the effectiveness of the proposed method and apply it to learn diverse and mixed reward functions of different human drivers from a naturalistic human driving dataset on a highway. The main contributions of this paper are listed as follows.

\begin{enumerate}
\item We apply maximum entropy inverse reinforcement learning with the proposed structural assumption to driving behavior modeling from naturalistic highway driving data. We show the effectiveness and efficiency of the proposed method in driving behavior modeling both qualitatively and quantitatively.

\item Two modeling assumptions, which are personalized modeling that each driver possesses a distinct cost function and general modeling that all drivers share a common cost function, are investigated. The personalized modeling method shows superiority over the general modeling method in terms of robustness and modeling accuracy.

\item The effects of simulating interactive behaviors in the environment model when evaluating the generated trajectories and incorporating the agent's interaction awareness into the reward function are investigated. The results indicate that the two factors are influential in estimating the reward of the generated trajectory and improving the modeling accuracy. The accuracy of applying the learned reward functions in the personalized planning process is also investigated.
\end{enumerate}


\section{Related Work}
\subsection{Driving behavior modeling}
There is a large body of literature on the topic of modeling human driving behavior, with a wide variety of problem formulations, model assumptions, and methodologies \cite{brown2020modeling}. The particular tasks of driving behavior modeling mainly fall into three groups: intention estimation, motion prediction, and pattern analysis. Intention estimation is to identify what a driver intends to do in the immediate future. Some classic methods include the parametric models such as the intelligent driver model (IDM) \cite{treiber2000congested} and minimizing overall braking induced by lane changes (MOBIL) model \cite{kesting2007general}, and the data-driven models like the hidden Markov model \cite{yuan2018lane}. For the motion prediction task that predicts the future physical states of a vehicle, neural network models are the dominant method, such as convolutional neural networks \cite{mo2020interaction} and long short-term memory networks \cite{8933492}. The parametric models generally postulate some structure about the problem, and thereby they are high interpretable and computationally efficient. However, the parametric models are not very expressive to reflect complex dynamics and the parameters are hard to specify. Data-driven methods do not make strong structural assumptions, instead, they rely on a wealth of data to extract the patterns underlying agent behaviors and make predictions. Such methods enjoy a strong performance but lack interpretability and generalization, which limits their application to safety-critical problems. In this work, we utilize the rough formulation of agents’ internal decision-making schemes, which is mathematically formulated as the reward function. It is a generalized form of decision-making processes and thus easier to integrate into many control and planning frameworks, and this work proposes to extract the parameters of the reward functions from naturalistic human driving data. 

On the other hand, pattern analysis is also an important branch in driving behavior research, which is to extract features or patterns from human driving data that can help us understand the traits of driving behaviors. For example, Siami \textit{et al.} developed an unsupervised pattern recognition framework to extract driving patterns from mobile telematics data, and they found 29 unique driving styles from the data \cite{siami2020mobile}. Birrell \textit{et al.} examined the correlation between certain parameters of human driving behavior and good fuel economy in real-world driving scenarios \cite{birrell2014analysis}. More recently, with the advent of the assisted driving system, researching driving behavior modeling in the driver-vehicle system at a micro level has gained great interest. For instance, Na and Cole proposed to utilize game theory to model a human driver's steering control behavior in response to vehicle automated steering intervention \cite{na2020theoretical}. Xing \textit{et al.} presented a deep learning-based joint driver behavior reasoning system to recognize both the driver's physical and mental states \cite{xing2020multi}. 

\subsection{Inverse reinforcement learning and its applications in driving behavior modeling}
Many models of human driving behavior employ an optimization setting, which postulates that human behavior is to optimize the expected reward of actions over time. Therefore, IRL has seen widespread use in many works as a tool to infer reward functions from expert demonstrations. The core idea of IRL is to adjust the weights of the reward function to yield a policy that matches the expert demonstrations (trajectories). Many IRL algorithms have been applied in driving behavior modeling, including the maximum margin method to learn driving styles and maneuver preferences \cite{silver2013learning} and the maximum entropy method to learn individual styles \cite{kuderer2015learning}. The maximum entropy method is more widely used as it can address the ambiguity that multiple reward functions can explain the expert's behavior. On the other hand, Wulfmeier \textit{et al.} proposed the deep maximum entropy IRL that can learn highly nonlinear cost map from raw high dimensional state input and applied it in large-scale vehicle navigation tasks \cite{wulfmeier2017large}. Although the network parameterization can be very expressive, the interpretability and generalization capability could be impaired. Since we want to explicitly represent and interpret human driving behavior, the maximum entropy method with linear reward function setting is more favorable.

The biggest challenges of maximum entropy IRL in driving behavior modeling are the continuous and large state spaces and computationally expensive RL process to evaluate the reward function at each iteration. To this end, many works choose to optimize a sequence of actions or a trajectory instead of stepwise actions in the evaluation process. The remaining challenge is the continuous and high dimensional state space, which makes it intractable to compute the partition function. \cite{sun2018probabilistic, hu2019generic} used Laplace approximation to reshape the reward function of a trajectory considering only local optimal \cite{levine2012continuous}, enabling the partition function to be solved analytically, but the assumption of local optimal may not stand in real-world cases. \cite{kuderer2015learning} proposed to optimize the spline trajectory with the updated reward function, and only the optimal trajectory is considered to calculate the feature expectation to update the reward function. Likewise, \cite{gonzalez2018modeling} utilized a spatiotemporal state lattice planner to search for an optimal trajectory, which is then used to calculate feature expectation in the IRL framework. However, the assumption of using a single optimal trajectory may be too strong while the human demonstrations can be sub-optimal and multi-optimal, and the optimization algorithms could be very time-consuming in optimizing long-horizon trajectories. 

Instead of directly optimizing a trajectory, another course is to sample trajectories, which can also be used to approximate the partition function. \cite{rosbach2019driving} proposed to sample a set of actions at each step in a planning cycle, resulting in a set of policies (trajectories) that encodes multiple behaviors. However, it only deals with static environments with only a vehicle dynamics model as the environment model, and thus more complex human driving behaviors from naturalistic driving are missing. \cite{xu2020learning} considered generating a set of trajectories rather than sampling low-level actions and learned the cost function by minimizing the discrepancy between expert and planned trajectories instead of the feature expectation in the general IRL setting. Our method closely relates to \cite{wu2020efficient}, where the authors suggested generating a trajectory set with elastic band path planning and speed profile sampler to estimate the partition function. However, these works are not focused on driving behavior modeling, thus lacking a structural assumption about human driving behavior that can facilitate reward learning as well. Moreover, the interaction behaviors of the surrounding agents in the environment are not well established in these works. To solve the problem of driving behavior modeling, we put forward a reasonable structural assumption on human driving behavior that can seamlessly integrate into the maximum IRL framework.

Additionally, one limitation of previous studies is the assumption that all vehicles in the human driving dataset share a common cost function, which certainly violates the fact that human driving behaviors are diverse and personalized. In real-world scenarios, human drivers can have distinct preferences, which entails learning reward functions about multiple intentions involving multiple agents \cite{babecs2011apprenticeship, ramponi2020truly}. In this paper, we focus on personalized driving behaviors with the assumption that each driver has a unique, personalized reward function.

\section{Methodology}
\label{sec2}
\subsection{Problem formulation}
Consider a human driver in an arbitrary traffic scene, the state $\mathbf{s}_t \in \mathcal{S} $ the driver observes at timestep $t$ consists of the positions, orientations, and velocities of itself and surrounding vehicles. The action $\mathbf{a}_t \in \mathcal{A}$ the driver takes is composed of speed and steering controls of the ego vehicle. Assuming a discrete-time setup and a finite length $L$, a trajectory $\zeta = \left[ \mathbf{s}_1, \mathbf{a}_1, \mathbf{s}_2, \mathbf{a}_2 \dots, \mathbf{s}_L, \mathbf{a}_L \right]$ is yielded by organizing the state and action in each timestep within the decision horizon. Note that the trajectory includes multiple vehicles in the driving scene since we consider interactions between agents. 

\begin{figure*}[htp]
    \centering
    \includegraphics[width=0.9\linewidth]{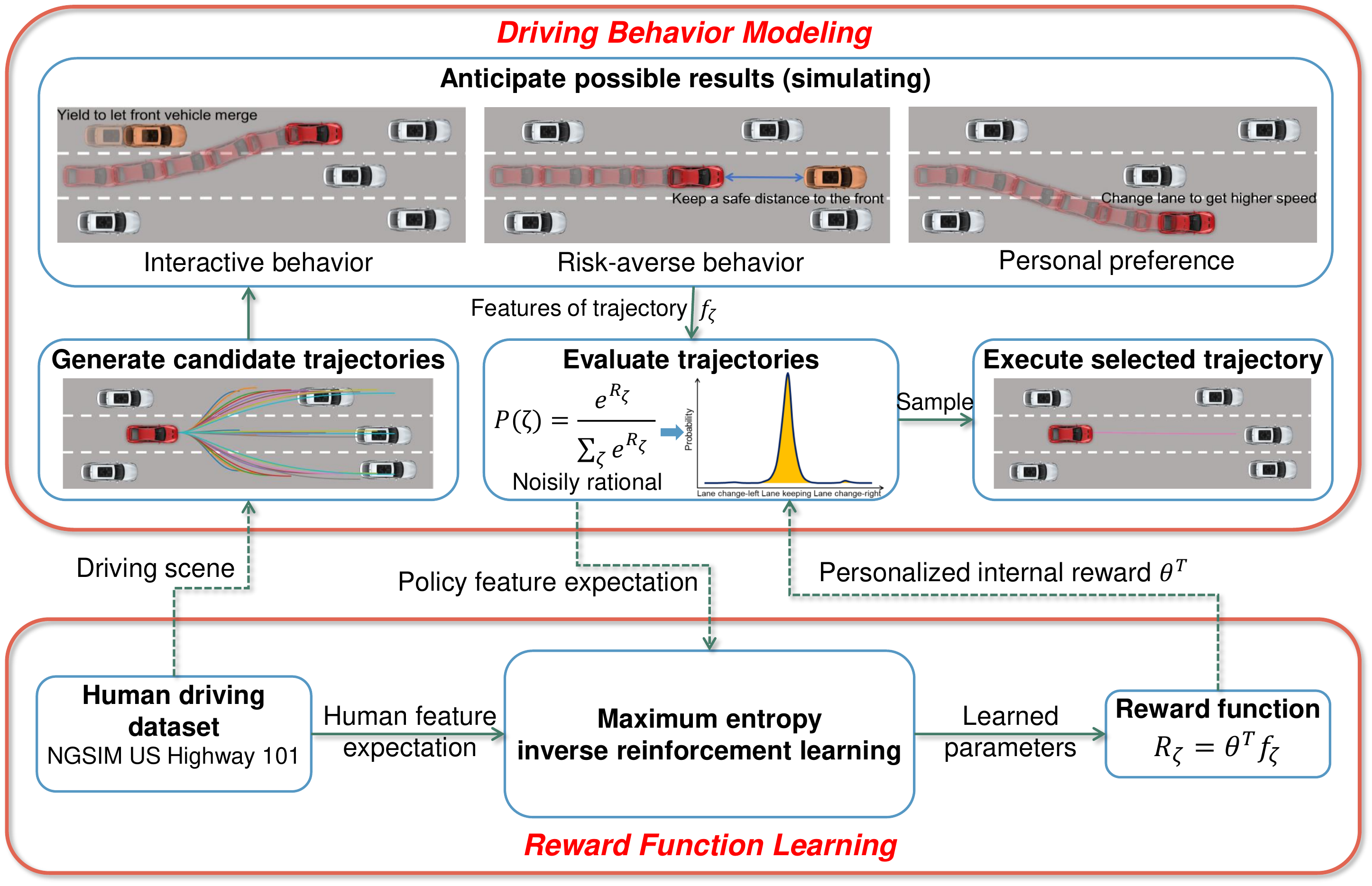}
    \caption{The framework of internal-reward-function-based driving behavior modeling with maximum entropy inverse reinforcement learning to infer the reward.}
    \label{fig1}
\end{figure*}

The state $\mathbf{s}_t$ is just a physical or partial observation that can be directly obtained from sensors. The latent intention, which may encompass attributes like the driver's navigational goals and driving intentions, actually governs the driver's actions. Based on this fact, we propose a structural assumption about human driving behavior that focuses on high-level intentions instead of low-level control actions, which is illustrated in Fig. \ref{fig1}. The assumption states that the human driving behavior consists of three processes, namely trajectory generation, trajectory evaluation, and trajectory selection. Given a driving scenario, a human driver first creates multiple candidate trajectories in mind, which relate to high-level decisions (e.g., lane-changing and lane-keeping) and speed requirements. At the same time, the driver should anticipate the results of the trajectories (involving interactive and risk-averse behaviors) and evaluate the returns of different trajectories with their internal reward functions (involving personal preference). The potential plans are assigned with probabilities according to the Boltzmann noisily-rational model (i.e., the probability of a trajectory is exponential to the reward of the trajectory), and finally the driver would execute one of the trajectories subjecting to the distribution. This assumption is justifiable and intuitive, and can well explain the stochasticity of human driving behavior. Besides, the probabilistic setting can address the suboptimal and multi-optimal policies existing in naturalistic human driving datasets.

We assume a linear-structured reward function, which is a weighted sum of the selected features. With a focus on the highway scenario with a simple road structure, where the driving pattern is relatively settled and the driver's preference or behavior would not change too much over time, we assume the weights of the reward function are consistent. Therefore, the reward function $r(\mathbf{s}_t)$ at a specific state $\mathbf{s}_t$ is defined as:
\begin{equation}
r(\mathbf{s}_t) = \boldsymbol \theta ^T \mathbf{f}(\mathbf{s}_t),    
\end{equation}
where $\boldsymbol \theta = \left[ \theta_1, \theta_2, \dots, \theta_K \right]$ is the $K$-dimensional weight vector and $\mathbf{f}(\mathbf{s}_t)$ is the extracted feature vector $\mathbf{f}(\mathbf{s}_t) = \left[ f_1(\mathbf{s}_t), f_2(\mathbf{s}_t), \dots, f_K(\mathbf{s}_t) \right]$ that characterizes the state $\mathbf{s}_t$. Therefore, the reward of a trajectory $R(\zeta)$ is given as:
\begin{equation}
R(\zeta) = \sum_{t} r(\mathbf{s}_t) = \boldsymbol \theta ^T \mathbf{f}_{\zeta} = \boldsymbol \theta ^T \sum_{\mathbf{s}_t \in \zeta}\mathbf{f}(\mathbf{s}_t),    
\end{equation}
where $\mathbf{f}_{\zeta}$ denotes the accumulative features along the trajectory $\zeta$. The selection of features is presented in Section \ref{sec4_B}.

Formally, given the human driving demonstration dataset $\mathcal{D} = \left\{ \zeta_1, \zeta_2, \dots, \zeta_N \right\}$ consisting of $N$ trajectories, the problem is to obtain the reward weights $\boldsymbol \theta$ that can generate a driving policy to match the human demonstration trajectories. We adopt the maximum entropy IRL algorithm to infer the reward weights $\boldsymbol \theta$ with the help of our proposed structural assumption on human driving behavior.

\subsection{Maximum entropy inverse reinforcement learning}
According to our assumption, a human driver follows a stochastic policy, which induces a distribution over generated candidate trajectories, and we assume the distribution is a Boltzmann distribution related to the returns of trajectories. This kind of distribution also has the maximum entropy among all such distributions that match the feature expectation of expert demonstrations, which corresponds to the maximum entropy IRL \cite{ziebart2008maximum, levine2012continuous}. Formally, the probability of a trajectory is proportional to the exponential of the reward of that trajectory, given by
\begin{equation}
P(\zeta|\boldsymbol \theta) = \frac{e^{R(\zeta)}}{Z(\boldsymbol \theta)} = \frac{e^{\boldsymbol \theta^T \mathbf{f}_\zeta}}{Z(\boldsymbol \theta)},
\end{equation}
where $P(\zeta|\boldsymbol \theta)$ is the probability of a trajectory $\zeta$ given reward parameter $\boldsymbol \theta$, and $Z(\boldsymbol \theta)$ is the partition function.

However, the partition function $Z(\boldsymbol \theta)$ is intractable for continuous and high dimensional spaces because it requires integrating over the entire class of possible trajectories. Referring to our assumption, the space of possible trajectories can be reduced to some small sub-spaces. Therefore, we generate a limited number of feasible trajectories, which are then used to approximate the partition function, and thus the probability of a trajectory becomes:
\begin{equation}
\label{eq4}
P(\zeta|\boldsymbol \theta) \approx \frac{e^{\boldsymbol \theta^T \mathbf{f}_{\zeta}}}{{\sum_{i=1}^M} e^{ \boldsymbol \theta^T \mathbf{f}_{\tilde{\zeta}^i}}},
\end{equation}
where $\tilde{\zeta}^i$ is a generated trajectory that has the same initial state as $\zeta$, $\mathbf{f}_{\tilde{\zeta}^i}$ the feature vector of the trajectory, and $M$ the number of generated trajectories. By doing this approximation, we make $P(\zeta|\boldsymbol \theta)$ a probability mass, which is much easier to compute.

The goal of maximum entropy IRL is to adjust the weights $\boldsymbol \theta$ to maximize the likelihood of expert demonstrations under the trajectory distribution in Eq. (\ref{eq4}):
\begin{equation}
\label{eq5}
\max_{\boldsymbol \theta} \mathcal{J} (\boldsymbol \theta) = \max_{\boldsymbol \theta} \sum_{\zeta \in \mathcal{D}} \log P(\zeta|\boldsymbol \theta),
\end{equation}
where $\mathcal{D}=\{ \zeta_i \}_{i=1}^N$ is the trajectory set of human demonstrations.

Substituting $P(\zeta|\boldsymbol \theta)$ in Eq. (\ref{eq5}) with Eq. (\ref{eq4}), we can rewrite the objective function $ \mathcal{J} (\boldsymbol \theta)$ as:
\begin{equation}
\label{eq6}
\mathcal{J} (\boldsymbol \theta) = \sum_{\zeta \in \mathcal{D}} \left[ \boldsymbol \theta^T \mathbf{f}_{\zeta} - \log {\sum_{i=1}^M} e^{ \boldsymbol \theta^T \mathbf{f}_{\tilde{\zeta}^i}} \right].
\end{equation}

Although cannot be solved analytically, Eq. (\ref{eq6}) can be optimized using a gradient-based method. The gradient of the objective function $\mathcal{J} (\boldsymbol \theta)$ is:
\begin{equation}
\label{eq7}
\nabla_{\boldsymbol \theta} \mathcal{J} (\boldsymbol \theta) = \sum_{\zeta \in \mathcal{D}} \left[ \mathbf{f}_{\zeta} - \sum_{i=1}^M \frac{e^{\boldsymbol \theta^T \mathbf{f}_{\tilde\zeta^i}}}{{\sum_{i=1}^M} e^{ \boldsymbol \theta^T \mathbf{f}_{\tilde{\zeta}^i}}} \mathbf{f}_{\tilde{\zeta}^i} \right],
\end{equation}
where $\mathbf{f}_{\zeta}$ is the feature vector of a human demonstrated trajectory, $\tilde{\zeta^i}$ is one of the generated trajectories that share the initial state of $\zeta$, and $\mathbf{f}_{\tilde\zeta^i}$ is the feature vector of that trajectory.

The gradient can be seen as the difference of feature expectations between the human demonstration trajectories and the generated ones:
\begin{equation}
\label{eq8}
\nabla_{\boldsymbol \theta} \mathcal{J}  (\boldsymbol \theta) = \sum_{\zeta \in \mathcal{D}} \left[ \mathbf{f}_{\zeta} - \sum_{i=1}^M P(\tilde\zeta^i|\boldsymbol \theta) \mathbf{f}_{\tilde{\zeta}^i} \right].
\end{equation}

We can use the gradient ascent method to iteratively update the reward function until the loss converges. In practice, to prevent overfitting, we add L2 regularization on the weights into the objective function:
\begin{equation}
\mathcal{J}(\boldsymbol \theta) = \sum_{\zeta \in \mathcal{D}} \left[ \boldsymbol \theta^T \mathbf{f}_{\zeta} - \log {\sum_{i=1}^M} e^{ \boldsymbol \theta^T \mathbf{f}_{\tilde{\zeta}^i}} \right] - \lambda \boldsymbol \theta^2,
\end{equation}
where $\lambda > 0$ is the regularization parameter. Thus, the gradient becomes the difference of the feature expectations plus the gradient of the regularization term:
\begin{equation}
\nabla_{\boldsymbol \theta} \mathcal{J}  (\boldsymbol \theta) = \sum_{\zeta \in \mathcal{D}} \left[ \mathbf{f}_{\zeta} - \sum_{i=1}^M P(\tilde\zeta^i|\boldsymbol \theta) \mathbf{f}_{\tilde{\zeta}^i} \right] - 2 \lambda \boldsymbol \theta.
\end{equation}

\subsection{Trajectory generation}
In order to efficiently generate feasible trajectories in a structured environment (highway in our case), we assume a human driver makes a short-term plan from the current state to an end target and considers the longitudinal and lateral targets respectively. For the longitudinal direction, the driver decides the target speed and for the lateral direction, and a tactical decision on lane-changing and lane-keeping is made for the lateral direction. Therefore, it is very convenient to use polynomial curves to represent the planned trajectories.

We use the local coordinate with reference to the origin and path of the road, to represent the trajectory of a vehicle into the longitudinal and lateral axis. The generated trajectory should be smooth and dynamically feasible, which requires the acceleration and jerk along the trajectory should be time-continuous. For lateral $y$-coordinate, we need to specify the target position, velocity, and acceleration, forming a total of six boundary conditions along with the initial state, which entails a quintic polynomial function. For longitudinal $x$-axis, only the target velocity and acceleration are needed, and thus a quartic polynomial function can fulfill the requirement of smoothness. Therefore, the trajectory is represented by two polynomial functions (continuous functions of time) with regard to $x$ and $y$ coordinate respectively:
\begin{equation}
\left\{\begin{matrix}
\mathbf{x}(\tau) = a_0 + a_1\tau + a_2\tau^2 + a_3\tau^3 + a_4\tau^4
\\
\mathbf{y}(\tau) = b_0 + b_1\tau + b_2\tau^2 + b_3\tau^3 + b_4\tau^4  + b_5\tau^5
\end{matrix}\right. ,
\end{equation}
where $\tau$ is the time, $\{a_0, \cdots, a_4\}$ and $\{b_0, \cdots, b_5\}$ are the coefficients of the polynomial functions.

Given the initial state of the ego vehicle and the target state, as well as the required time $T$ to reach the target, the boundary conditions of the polynomial functions on the longitudinal and lateral axis are:
\begin{equation}
\left\{ \begin{array}{l}
{\bf{x}}(\tau = 0) = {x_s}\\
{\bf{\dot x}}(\tau = 0) = {v_{xs}}\\
{\bf{\ddot x}}(\tau = 0) = {a_{xs}}\\
{\bf{\dot x}}(\tau = T) = {v_{xe}}\\
{\bf{\ddot x}}(\tau = T) = {a_{xe}}
\end{array} \right., \ 
\left\{ \begin{array}{l}
{\bf{y}}(\tau = 0) = {y_s}\\
{\bf{\dot y}}(\tau = 0) = {v_{ys}}\\
{\bf{\ddot y}}(\tau = 0) = {a_{ys}}\\
{\bf{y}}(\tau = T) = {y_e}\\
{\bf{\dot y}}(\tau = T) = {v_{ye}}\\
{\bf{\ddot y}}(\tau = T) = {a_{ye}}
\end{array} \right.
\end{equation}
where $(x_s, v_{xs}, a_{xs}, y_s, v_{ys}, a_{ys})$ is the start state ($\tau=0$) including position, velocity, and acceleration in the longitudinal and lateral directions; $(v_{xe}, a_{xe}, y_e, v_{ye}, a_{ye})$ is the target state ($\tau=T$) without the target longitudinal position.

By solving the boundary equations, the coefficients of the polynomial functions can be determined, and thereby a trajectory is generated. The position of the ego vehicle on the trajectory at any given time $\tau \ (\tau \le T$) can be derived, as well as the velocity and acceleration. In this paper, we set the horizon to 5 seconds ($T=5\ s$). Given the instants (with a time interval of 0.1 seconds) at which the acceleration, velocity, and position values are computed, a polynomial trajectory can be generated, which is a sequence of these values aligning with the timesteps of the human driving trajectories. 

\begin{figure}[htp]
    \centering
    \includegraphics[width=\linewidth]{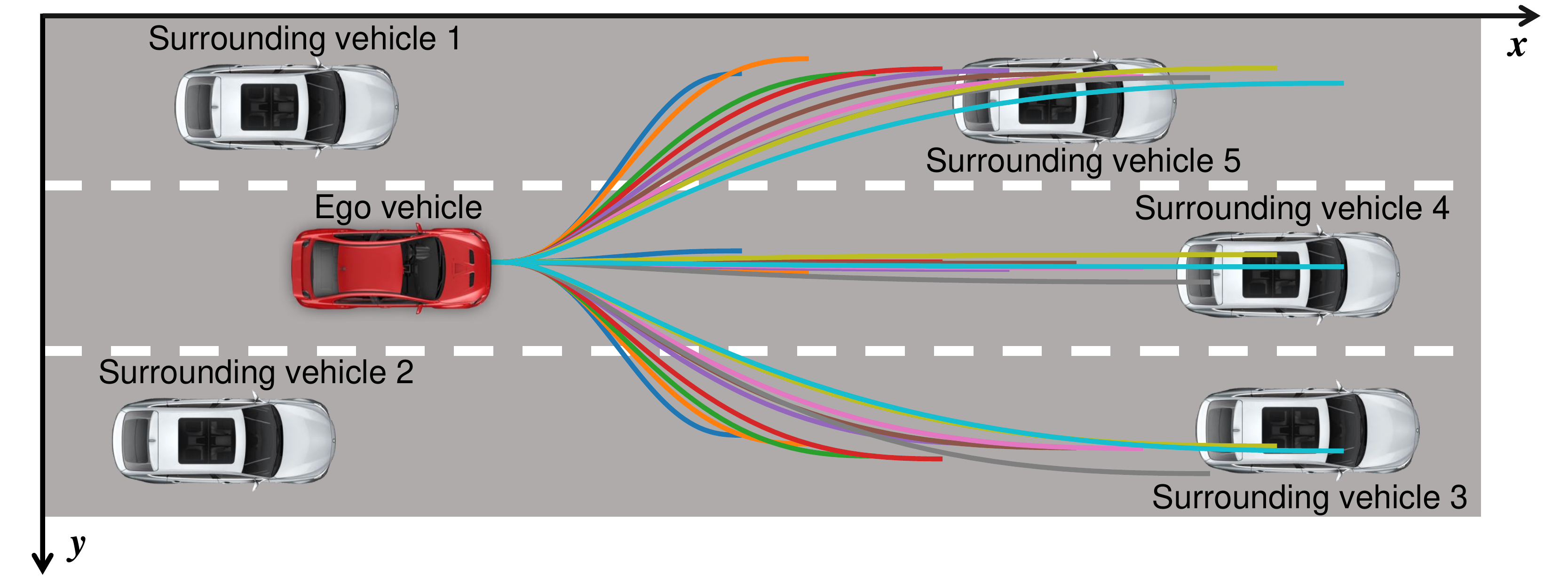}
    \caption{Trajectory generation process considering the longitudinal and lateral targets}
    \label{fig2}
\end{figure}

We can generate multiple polynomial trajectories by sampling the target state from the target space $\mathbf{\Phi}=\{ v_{xe}, a_{xe}, y_e, v_{ye}, a_{ye} \}$, which sufficiently covers possible maneuvers. Fig. \ref{fig2} shows an example of the trajectory generation process, in which only $v_{xe}$ and $y_e$ are variable while others are constant to 0. In Fig. \ref{fig2}, multiple candidate trajectories are generated covering the decisions on lane-changing and lane-keeping, as well as the desired longitudinal speed.

\subsection{Environment model}
To simulate the outcomes of the generated trajectories and the reactions of other agents to the generated actions of the ego, particularly those that notably deviate from the ground truth, an environment model is necessary. The model serves, in a sense, as a simulated mental world of human drivers, to anticipate other agents' reactions to the ego movements and help estimate the reward of the generated trajectory. We first construct the multi-lane highway road with the same structure as the study area in the NGSIM US-101 dataset, which consists of five mainline lanes throughout the section and an auxiliary lane between the on-ramp and the off-ramp. More details about the road structure can be found in Section \ref{sec4_A}. Then we spawn vehicles on the road as recorded in the dataset at a specific instant, and one of the vehicles is targeted as the observation object and maneuvered by a pure-pursuit controller to track the generated trajectory to guarantee the final trajectory is dynamically feasible. The kinematic state of the vehicle is propagated according to the bicycle model.

\begin{figure}[htp]
    \centering
    
    \subfloat[]{\includegraphics[width=0.95\linewidth]{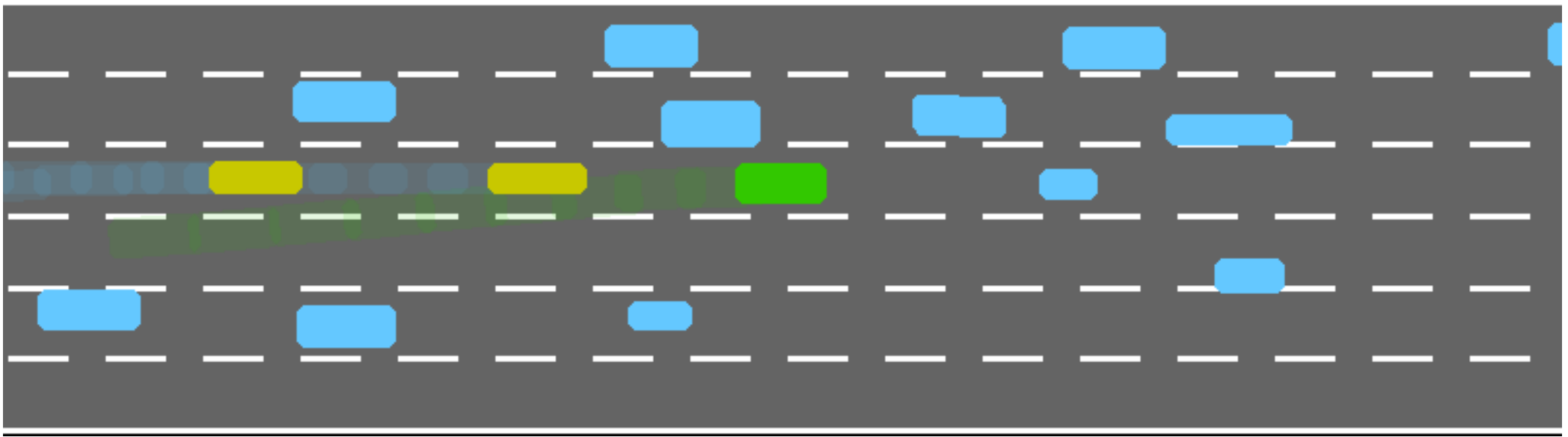}%
    \label{fig3a}}
    
    \subfloat[]{\includegraphics[width=0.95\linewidth]{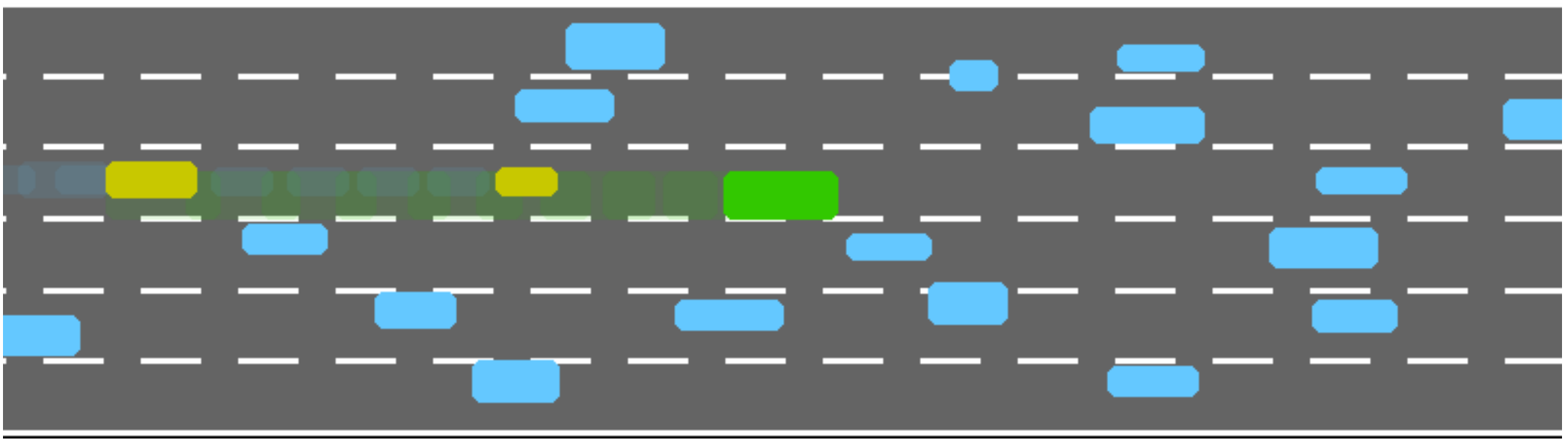}%
    \label{fig3b}}
    
    \subfloat[]{\includegraphics[width=0.95\linewidth]{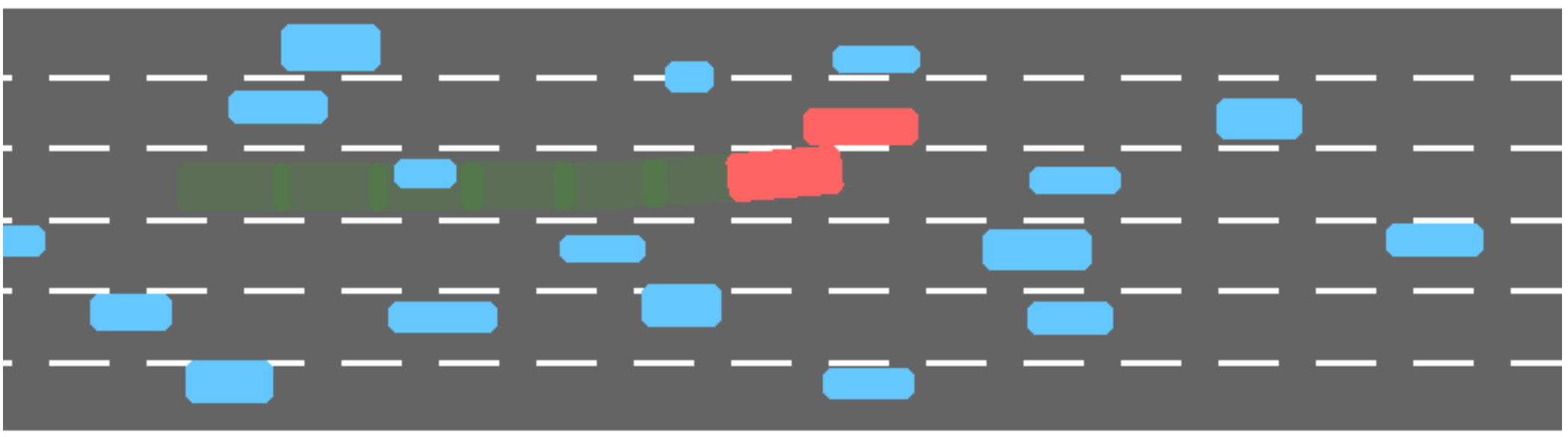}%
    \label{fig3c}}
    
    \caption{Illustrations of some interactive behaviors in the environment model: (a) the ego vehicle tries to change to the right lane, which makes the affected vehicles decelerate to yield; (b) the ego vehicle runs too slow, which makes the rear vehicles decelerate to avoid a collision; (c) the ego vehicle's generated trajectory causes a collision.}
    \label{fig3}
\end{figure}

Next, we need to predict the future trajectories of the surrounding vehicles in response to the ego vehicle's actions. The general idea is that the surrounding vehicles follow their original trajectories in the dataset and otherwise react by keeping a safe distance to the ego vehicle or the influenced vehicle. The underlying assumption is that humans are best-response agents who can anticipate other agents' reactions to their planned actions accurately. It is worth noting that this setting may introduce some bias in the estimation of the transition function. In detail, only the vehicles within a range of 50 meters to the ego vehicle in the environment are considered. Surrounding vehicles will come after their recorded trajectories at first and each of them constantly checks the gap between itself and the vehicles in the front. If the front is the ego vehicle and the gap between them is smaller than the desired gap given by IDM, the environment vehicle will be overridden by IDM and thus not follow its original trajectory anymore. Likewise, if the front vehicle is an environment vehicle that has been overridden by IDM, the environment vehicle behind will also be overridden if the gap between them is too small. IDM is a parametric car-following model for highway traffic simulation, which models the driver's desire to achieve the target speed and the need to maintain a safe distance with the vehicle in front. The inputs to the model are the vehicle’s current speed, the relative speed with respect to the leading vehicle, and distance headway, and the output of the model is the acceleration, and speed and position can be inferred subsequently. The model requires several parameters, such as the desired speed and minimum desired spacing, to represent different driving behaviors, and more details about IDM can be found in \cite{treiber2000congested}. We only consider the longitudinal responses of the surrounding vehicles to simulate the influence of the change-of-course behavior of the ego vehicle and coded by these simple rules, multiple surrounding vehicles can be affected sequentially by the ego vehicle's decisions. Fig. \ref{fig3} shows some exemplar scenarios where the vehicle in green is the ego vehicle, and the vehicles in blue are surrounding vehicles following their original trajectories. The yellow ones are the vehicles that have been affected by the ego vehicle's actions and thus been overridden by IDM. Vehicles becoming red indicate that they have collided with others. 

\subsection{Summary of the IRL algorithm}
The algorithm of maximum entropy IRL with trajectory sampling is summarized in Algorithm \ref{algo1}. We first initialize the reward parameters randomly and compute the feature expectation of human driving trajectories. Since the sampling process consumes most of the computation time, we create a buffer to store the feature vectors of all the generated trajectories to avoid iterative sampling in the environment. For each driving scene provided by the demonstration data, we generate a set of trajectories and roll them out in the environment model to get the feature vectors. The size of generated trajectory set $\tilde{\mathcal{D}}_i$ for one driving scene is determined by the size of target sampling space (i.e., the number of the longitudinal targets times the number of the lateral targets). After we have finished the sampling process and obtained the buffer, we can calculate the gradient and use the gradient ascent method to iteratively update the reward parameters, making the feature expectation of the generated trajectories match that of the human trajectories.

\begin{algorithm}[htp]
\label{algo1}
\SetAlgoLined
\SetKwInOut{Input}{Input}\SetKwInOut{Output}{Output}

\Input{Human demonstration trajectory dataset $\mathcal{D}=\{ \zeta_i \}_{i=1}^N$, environment model $P$, learning rate $\alpha$, regularization parameter $\lambda$, number of epochs $E$}
\Output{Optimized reward function parameters $\boldsymbol \theta^*$}
\BlankLine

Initialize $\boldsymbol \theta \leftarrow \mathcal{N}(0, 0.05)$\;
Compute human feature expectation $\mathbf{\bar{f}} \leftarrow \sum_{i=1}^N \mathbf{f}_{\zeta_i}$\;
Initialize buffer $\mathcal{B} \leftarrow [ \ ]$\;

\ForEach{$\zeta_i$ in $\mathcal{D}$}{
Determine the sampling space $\mathbf{\Phi}$ and planning horizon $T$\;
Generate a trajectory set $\mathcal{\tilde{D}}_i= \{ \tilde{\zeta}_i^j \}$ with the same initial state as $\zeta_i$ according to the sampling space $\mathbf{\Phi}$ and horizon $T$\;
    \ForEach{$\tilde{\zeta}_i^j$ in $\mathcal{\tilde{D}}_i$}{
    Rollout the trajectory $\tilde{\zeta}_i^j$ in the environment model $P$ and calculate the feature vector of the trajectory $\mathbf{f}_{\tilde{\zeta}_i^j}$ \;
    Add trajectory and its feature vector to buffer $\mathcal{B} \overset{+}{\leftarrow} \tilde{\zeta}_i^j, \ \mathbf{f}_{\tilde{\zeta}_i^j}$\;
    }
}

\For{epoch $\leftarrow$ 1 to E}{
Calculate the feature expectation with the collected samples from $\mathcal{B}$: $\mathbf{\tilde{f}} \leftarrow \sum_{i=1}^N \sum_j \frac{\exp \left( {\boldsymbol \theta^T \mathbf{f}_{\tilde{\zeta}_i^j}}\right)}{\sum_j \exp \left({\boldsymbol \theta^T \mathbf{f}_{\tilde{\zeta}_i^j}} \right)} \mathbf{f}_{\tilde{\zeta}_i^j}$\;
Calculate the gradient $\nabla_{\boldsymbol \theta} \mathcal{J}(\boldsymbol \theta) \leftarrow \mathbf{\bar{f}} - \mathbf{\tilde{f}} - 2 \lambda \boldsymbol \theta$\;
Update reward parameters $\boldsymbol \theta \leftarrow \boldsymbol \theta + \alpha \nabla_{\boldsymbol \theta} \mathcal{J}(\boldsymbol \theta)$\;
}
$\boldsymbol \theta^* \leftarrow \boldsymbol \theta$

\caption{Maximum entropy inverse reinforcement learning with trajectory sampling}
\end{algorithm}

\section{Experimental Validation}
\label{sec4}
\subsection{Naturalistic human driving dataset}
\label{sec4_A}
\begin{figure}[htp]
    \centering
    \includegraphics[width=\linewidth]{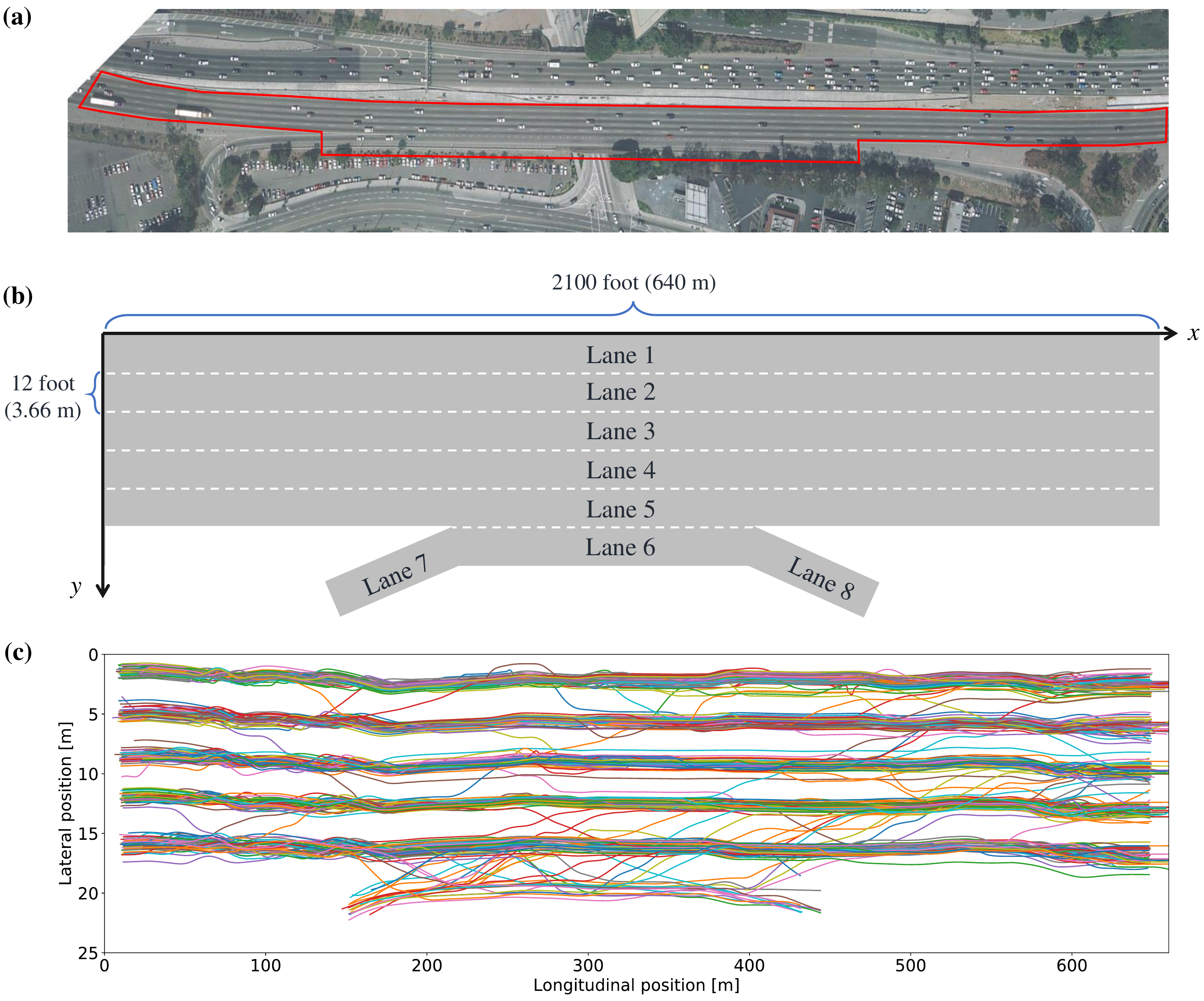}
    \caption{Illustration of the dataset and road structure: (a) the study area on the US Highway 101; (b) the reconstructed road structure with the local coordinate system; (c) the trajectories of randomly selected 300 vehicles in local coordinates.}
    \label{fig:4}
\end{figure}

To validate the proposed method for driving behavior modeling, we employ the Next Generation Simulation (NGSIM) dataset \cite{alexiadis2004next} with a segment of data within 7:50 a.m to 8:05 a.m. on the US Highway 101. The recording area is a section of the highway with approximately 640 meters (2,100 foot) in length and consists of five main lanes throughout the section and an auxiliary lane between an on-ramp and an off-ramp, as shown in Fig. \ref{fig:4}(a). In addition to the global positions, the dataset also provides the local positions of vehicles with respect to the local coordinate system. Based on the data, we reconstruct the road structure in our environment model, as shown in Fig. \ref{fig:4}(b). The origin of the coordinate system is on the top-left corner vertex of the study area. The longitudinal $x$ axis extends along the road while the lateral $y$ axis is perpendicular to the direction of the road. The length of the road is 640 meters with five main lanes (Lane 1 to Lane 5), each with a width of 3.66 meters. The on-ramp (Lane 7) and off-ramp (Lane 8) and the auxiliary lane between them (Lane 6) are also considered but the ramp merging scenarios are not the focus of this paper. The locations of each vehicle are recorded at 10 frames per second, resulting in detailed vehicle trajectories, from the start of the area to the end. However, the originally collected vehicle trajectories in the dataset are full of observation noise, and thus we use the Savitzky-Golay filter with a third-order polynomial over 2-second windows to smooth the original trajectories and obtain the demonstration trajectories for reward learning. Fig. \ref{fig:4}(c) shows the processed trajectories of randomly selected 300 vehicles in the dataset as examples, and the speeds and accelerations can be estimated from position data. The naturalistic human driving dataset encompasses the trajectories of nearly 3000 vehicles and provides rich information on interactions between human drivers, which is necessary for our study to reveal the diverse, personalized, and highly interactive human driving behaviors.

\subsection{Feature selection}
\label{sec4_B}
Features are mappings from state to real values which capture important properties of the state. Here, we group the features of the driving state in the following four main aspects that are important to human drivers \cite{naumann2020analyzing}. 

\subsubsection{Travel efficiency} 
this feature is designed to reflect the human's desire to reach the destination as fast as possible, which is defined as the speed of the vehicle:
\begin{equation}
f_{v}(\mathbf{s}_t) = v(t). 
\end{equation}

\subsubsection{Comfort}
ride comfort is another factor that human drivers prefer, and the metrics to gauge comfort are longitudinal acceleration $a_{x}$, lateral acceleration $a_{y}$, and longitudinal jerk $j_{x}$:
\begin{equation}
\left\{\begin{matrix}
f_{a_{x}}(\mathbf{s}_t) = \left| a_x(t) \right| = \left| \ddot{x}(t) \right|
\\
f_{a_{y}}(\mathbf{s}_t) = \left| a_y(t) \right| = \left| \ddot{y}(t) \right|
\\
f_{j_{x}}(\mathbf{s}_t) = \left| \dot{a}_x(t) \right| = \left| \dddot{x}(t) \right|
\end{matrix}\right.,
\end{equation}
where $x(t)$ and $y(t)$ are the longitudinal and lateral coordinates, respectively.

\subsubsection{Risk aversion}
a human driver tends to keep a safe distance to the surrounding vehicles and this distance varies across different human drivers, which reflects their different levels of sensing risk. We define the risk level to the front vehicle as an exponential function related to the time headway from the ego vehicle to the front vehicle assuming a constant speed movement:
\begin{equation}
f_{risk_f}(\mathbf{s}_t) = e^{ - \left( \frac{x_f(t) - x_{ego}(t)}{v_{ego}(t)} \right)},
\end{equation}
where $x_f(t)$ is the longitudinal position of the nearest front vehicle, $x_{ego}(t)$ is that of the ego vehicle, and $v_{ego}(t)$ is the speed of the ego vehicle.

Likewise, the risk level to the rear end is defined as an exponential function related to the time headway from the rear vehicle to the ego vehicle:
\begin{equation}
f_{risk_r}(\mathbf{s}_t) = e^{ - \left( \frac{x_{ego}(t) - x_r(t)}{v_{r}(t)} \right)},
\end{equation}
where $x_r(t)$ and $v_r(t)$ are the longitudinal position and speed of the nearest rear vehicle, respectively.

Note that collision may happen when evaluating the generated trajectories in our environment model, including colliding with other vehicles or road curbs, so the collision is also a risk indicator, which is defined as:
\begin{equation}
f_{collision}(\mathbf{s}_t) = \begin{cases}1 & \text{if collision}, \\0 & \text{otherwise}.\end{cases}
\end{equation}


\subsubsection{Interaction}
a fundamental property of human driving behavior is that humans are aware of the influence of their actions on the surrounding vehicles, or more specifically if their plans would impose additional inconvenience to other people (e.g., sharp decelerate to yield) \cite{evestedt2016interaction}. We introduce the following feature to explicitly represent such influence. It is defined as the sum of predicted decelerations of the environment vehicles that have been affected by the behavior of the ego vehicle according to our environment model (chain deceleration reactions of the following vehicles), indicating that the ego vehicle's change of the original course has caused direct influence on them.

\begin{equation}
\label{eq16}
f_{I}(\mathbf{s}_t) = \sum_i | a_i (t) |, \ \text{if} \ a_i (t) < 0,
\end{equation}
where $a_i(t)$ is the acceleration of the vehicle $i$ that has been influenced by the ego vehicle. When it comes to applying the reward function in real-world scenarios, we can estimate this feature with a prediction module that forecasts other agents' actions due to the planned actions using driving models such as IDM.

All the above features are calculated at every timestep and accumulated over time to obtain the feature of the trajectory. The trajectory features are then normalized to $[0,1]$ by dividing by the maximum value in the dataset to cancel out the influence of their different units and scales. Additionally, we assign a fixed large negative weight (-10) on the collision feature because this could improve the modeling accuracy than making this weight learnable.

\subsection{Experiment design}
\subsubsection{Driving behavior analysis}
we utilize the proposed method to analyze the driving behaviors of different human drivers. We first show the reward learning process of a human driver from the dataset as an example to reveal the effectiveness of our proposed method. Then, the learned reward function is used to determine the probabilities of the candidate trajectories in testing conditions and interpret some driving behaviors.

\subsubsection{Robustness}
we test the learned reward functions in the scenes that are not in the training phase to find out if there is a significant drop in the similarity between the learned and human policies, in order to investigate the robustness of the proposed method.

\subsubsection{Modeling accuracy}
we show the quantitative results of modeling accuracy in the testing conditions by comparing the learned policy to the ground-truth human driving trajectory. We investigate the personalized modeling assumption that each human driver has different preferences (driving styles), thus having different weights over the reward function. The general modeling assumption that all drivers share an identical cost function is adopted as a comparison. Two other baseline models are also employed, which are IDM and MOBIL for longitudinal and lateral movement respectively, and the constant velocity model.

\subsubsection{Interaction factors}
we analyze the effects of interaction factors on the modeling accuracy. They include the interaction feature in the reward function and simulating surrounding vehicles' reactions to the change of course of the ego vehicle in the environment model. 


\subsection{Implementation details}
For simplicity, the target sampling space is reduced to $\mathbf{\Phi}=\{ v_{xe}, y_e \}$, in which only the longitudinal speed and lateral position are variables and other targets are set as 0. The sampling range of the longitudinal speed is $\left[v-5, v+5 \right]$ m/s with an interval of 1 m/s, where $v$ is the initial speed of the vehicle. The sampling set of the lateral position is $\{ y, y_L, y_R \}$ m, where $y$ is the initial lateral position, and $y_L$ and $y_R$ are the position of the left lane and right lane, respectively, if they are available. The planning horizon is $5 \ s$ and the simulation interval is $0.1 \ s$. The parameters of IDM are: desired velocity $v_0 = v_{current} \ m/s$, maximum acceleration $a_{max} = 5 \ m/s^2$, desired time gap $\tau = 1 \ s$, comfortable braking deceleration $b = 3 \ m/s^2$, minimum distance $s_0 = 1 \ m$. A problem emerges that the longitudinal and lateral jerk of human trajectories and generated trajectories can hardly match because the polynomial curves are smooth while the human driving trajectories are full of noisy movements. Therefore, we process the human driving trajectory to be represented by a polynomial curve given the initial state and end condition of the original trajectory. 

To stabilize the training process, Adam optimizer instead of the vanilla gradient ascent method is implemented. There are three hyperparameters that need to be tuned, which are the regularization parameter $\lambda$, the learning rate $\alpha$, and the number of training epochs $E$. We use the grid search method with the parameter space as $\lambda \in \{0.1, 0.01, 0.001\}$, $\alpha \in \{0.1, 0.05, 0.01\}$, and $E \in \{100, 200, 300\}$, and the performance metric as the average likelihood of demonstration trajectories from 10 drivers. The final setting of the hyperparameters is $\lambda=0.01, \alpha=0.05, E=200$. The time it takes to finish the learning process (Algorithm \ref{algo1}) with 35 human demonstration trajectories from a vehicle is roughly 18 minutes with an AMD Ryzen 3900X CPU. In the testing phase, it takes about 30 seconds to implement a sampling and evaluation process (with the size of sampled trajectories as 30). Note that parallel sampling is not used as real-time planning is not the focus of this paper but it can significantly speed up the learning and testing processes.

\section{Results and Discussions}
\label{sec4}
\subsection{Driving behavior analysis}
\begin{figure}[htp]
    \centering
    \includegraphics[width=\linewidth]{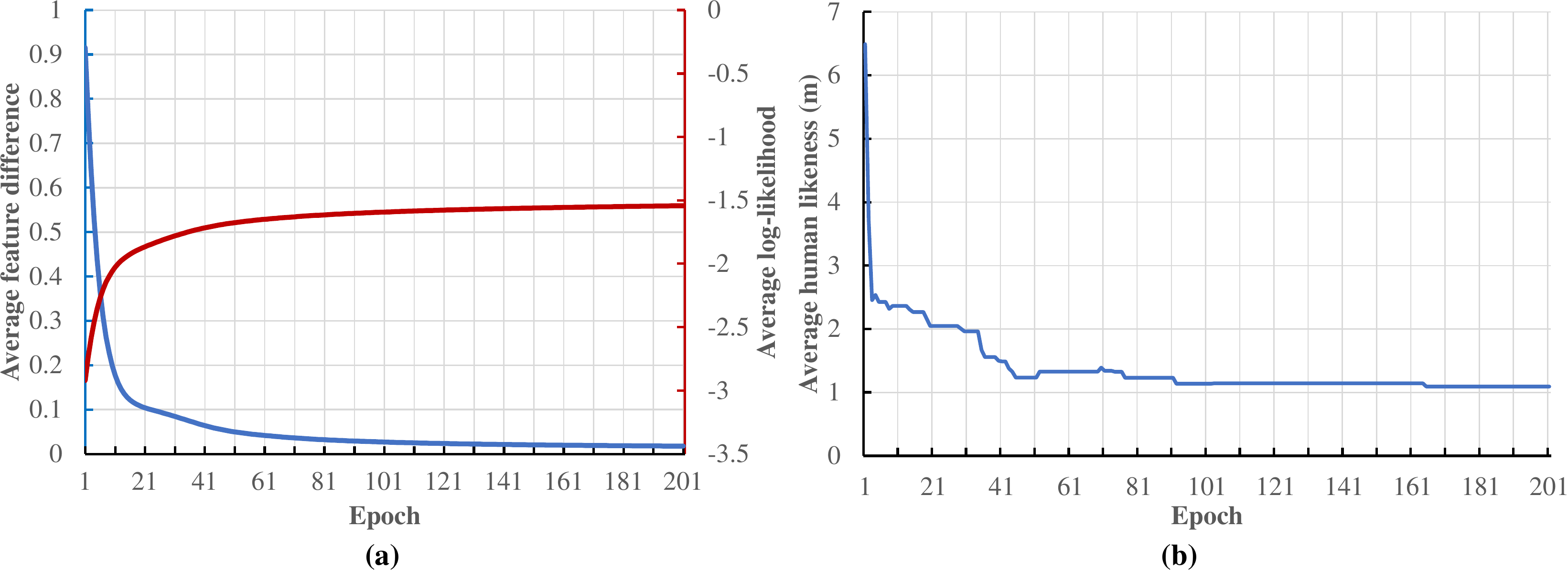}
    \caption{Example of the training process: (a) plot of the average feature difference and log-likelihood; (b) plot of the average human likeness.}
    \label{fig5}
\end{figure}

For a vehicle in the dataset, its original trajectory throughout the highway section, which is approximately 50 to 70 seconds in time length, is evenly partitioned into 50 short-term trajectories, each with $5 \ s$ length of time. Each trajectory represents a driving scene involving different situations and different kinds of interactions with the surrounding vehicles. 35 trajectories among them are randomly selected and serve as the training data for reward function learning. The rest 15 trajectories serve as the testing conditions, where the learned reward function is used to select the candidate trajectories. An example of the training process is shown in Fig. \ref{fig5}, which plots the curves of average feature difference (L2 norm) between the learned policy and human driver, average log-likelihood of the human demonstrated trajectories, and average human likeness. The human likeness is a custom metric to gauge the accuracy of the model, i.e., closeness to the ground-truth human driving behavior. Since our model is a probabilistic model, we define the human likeness as the minimal final displacement error of three trajectories with the highest probabilities in the distribution over generated trajectories. It is defined as the L2 norm between the position at the end of the ground truth trajectory and that of the closest prediction among the three most likely trajectories. Formally, $HL = \min \{ \big\| \hat{\zeta}_{i}(L)- \zeta_{gt}(L) \big\|_2 \}_{i=1}^3$, where $\hat{\zeta}_{i} \ (i=1,2,3)$ are the selected trajectories with the highest probabilities, $\zeta_{gt}$ is the ground-truth trajectory by the human driver, and $L$ is the end of the time horizon. Therefore, smaller human likeness means better modeling accuracy.

As seen in Fig. \ref{fig5}(a), the average log-likelihood of the human demonstrated trajectories gradually increases and converges, recalling that the goal of the maximum entropy IRL is to maximize the likelihood of human demonstrations, and the average feature difference between the learned policy and human driver steadily reduces to a small number. This gives rise to the decrease in human likeness as shown in Fig. \ref{fig5}(b), which means that the probability of choosing the trajectories close to human driving behavior raises under the learned reward function. The results justify both the effectiveness of the proposed maximum entropy IRL algorithm with trajectory sampling.


\begin{figure*}[htp]
    \centering
    \includegraphics[width=0.9\linewidth]{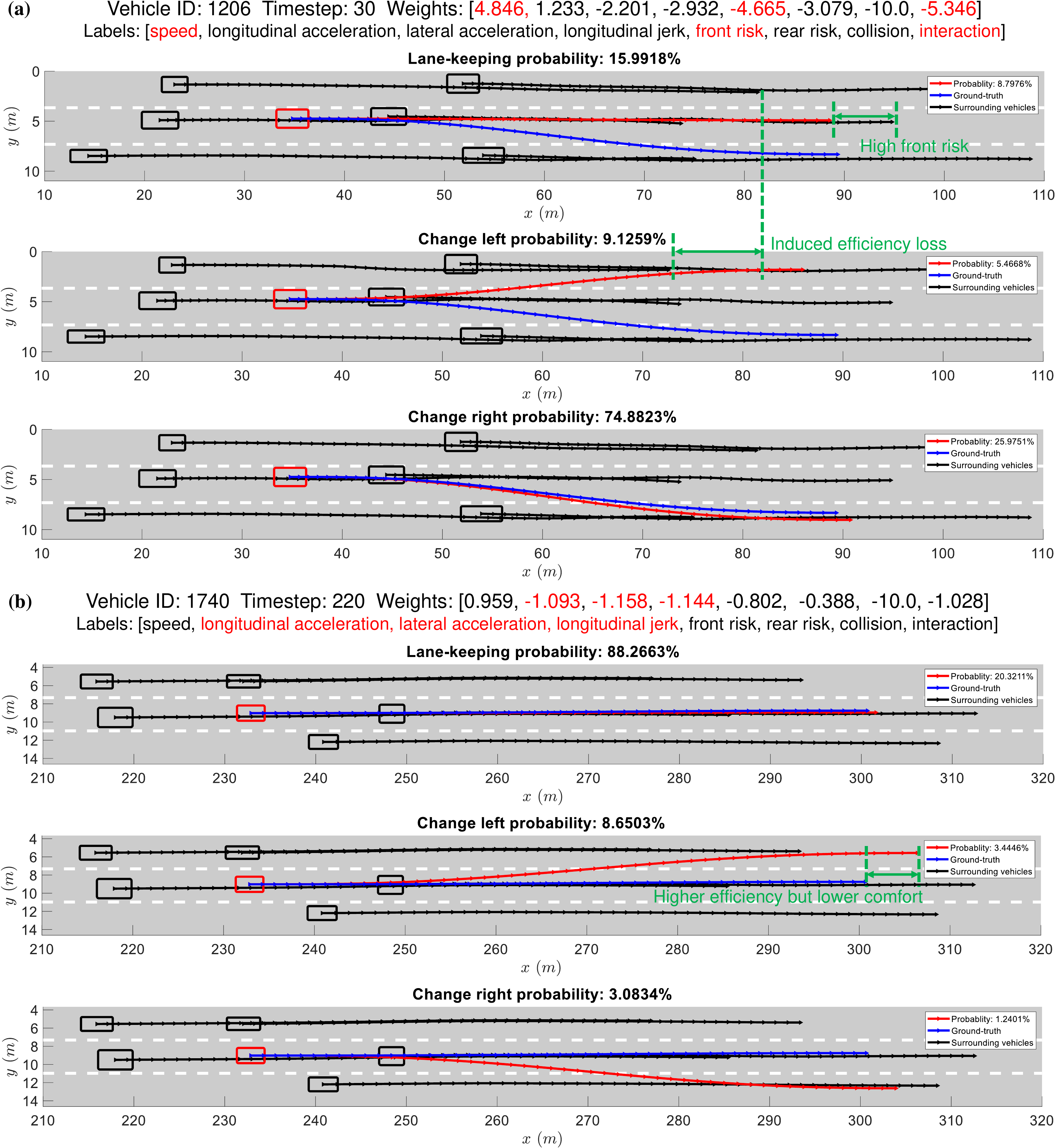}
    \caption{Driving behavior analysis of some typical cases from the US 101 highway dataset. The top-3 important features with higher weights are marked in red, except for the collision feature.}
    \label{fig6}
\end{figure*}

Then, we select various human drivers and associated driving trajectories from the dataset and apply the proposed method to infer their individual reward functions, and eventually use the learned reward to interpret their decisions. Fig. \ref{fig6} shows some representative cases of different vehicles from the US-101 highway dataset. The candidate trajectories and their associated probabilities and the ground-truth human driving trajectories are displayed, as well as the trajectories of the surrounding vehicles as the interaction context. Only the most likely trajectory in the three discrete lateral decision groups (lane-keeping, change left, and change right), is displayed in Fig. \ref{fig6}. Generally, lowering the risk to both the front end and rear end is a critical factor shared by most human drivers, while the other factors (speed, ride comfort, and interaction) varies among different human drivers. Fig. \ref{fig6}(a) shows an overtaking scenario, in which the human driver, as represented by the recovered reward function, views that the speed weighs more than the ride comfort (both longitudinally and laterally). If the driver stays in the current lane and wants to keep the speed, the distance to the front vehicle would be shorter, which is not likely to happen since it would bring higher front risk. Besides, if the driver chooses to change to the left lane, a notable speed loss can be imposed on the rear vehicle, which is an undesired result since the driver opposes imposing influence on others, given a higher weight on the interaction term. Therefore, the driver would choose to change to the right lane to overtake, as predicted by our model with a probability of nearly 75\%. A detailed trajectory is also given, which is highly close to the ground-truth human driving one. This example signifies that our model can accurately predict the lane-changing behavior and also produces detailed trajectories. For another instance, in Fig. \ref{fig6}(b), where the human driver treats the ride comfort (both longitudinally and laterally) as the main concern, even if changing to the left lane can bring higher efficiency, the driver would still keep the current lane to avoid acceleration and jerk, as predicted by the model with a probability of 88\%.

\subsection{Testing of accuracy and robustness}
\begin{figure}[htp]
    \centering
    \includegraphics[width=\linewidth]{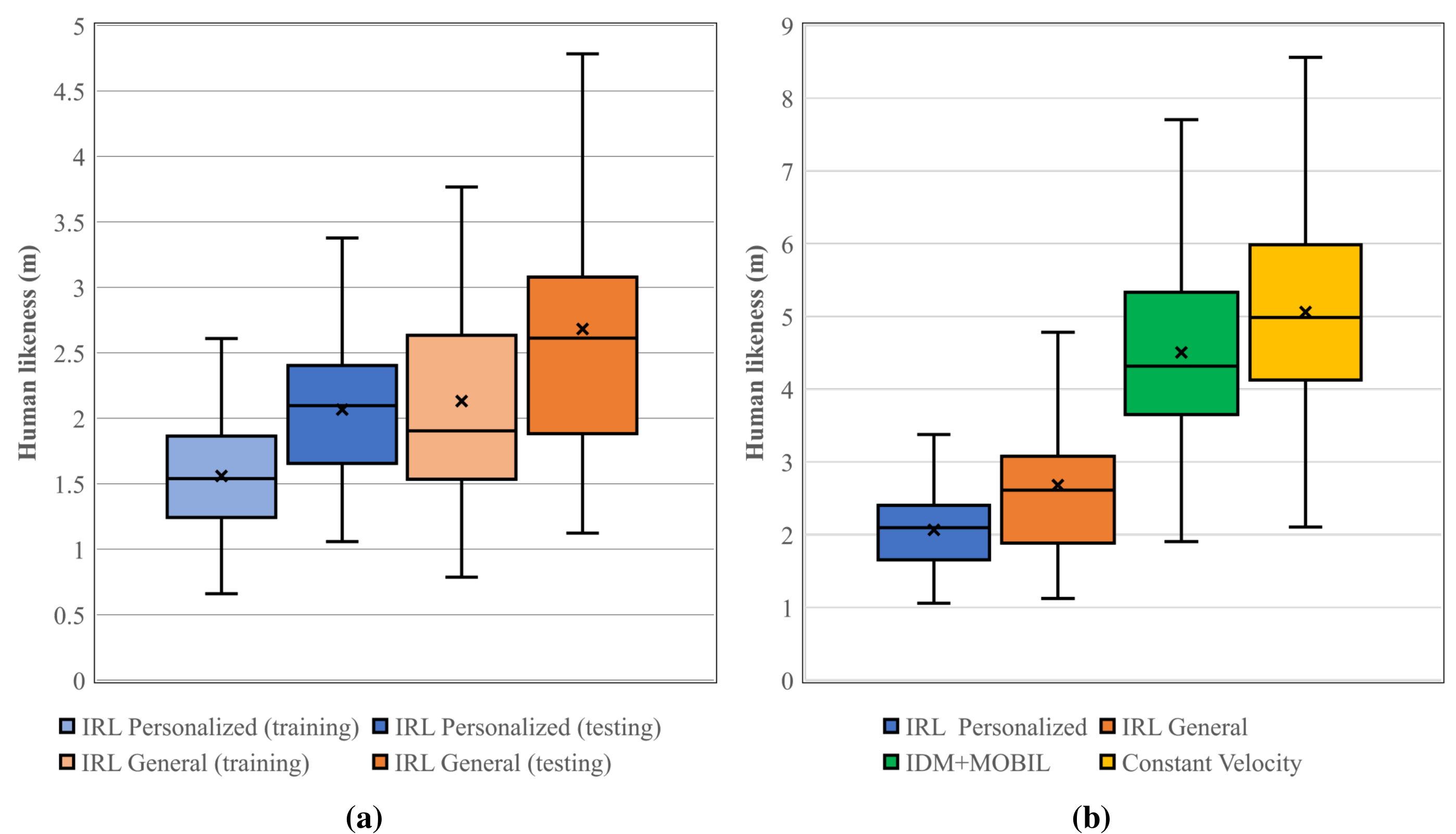}
    \caption{Comparison of robustness and modeling accuracy of different models: (a) robustness testing; (b) modeling accuracy testing.}
    \label{fig7}
\end{figure}

We randomly select 100 vehicles from the dataset, among which 50 vehicles experience lane change while the rest only run lane-keeping, as the target objects. For the personalized modeling assumption that each driver has a unique cost function, the proposed IRL method is applied to infer their individual reward functions. For each individual vehicle, we exam the robustness of the learned reward function in the on-hold 15 driving scenes different from the training conditions. For the general modeling assumption, a total of 150 trajectories from 20 vehicles are used to learn a general cost function, which is assumed to be shared by all human drivers. The learned cost function is utilized to select the candidate trajectories in the testing conditions same as the personalized modeling method and compared against the ground-truth human driving trajectories, so as to investigate the robustness of the learned reward function. The results of the robustness testing are given in Fig. \ref{fig7}(a). To reflect the accuracy of the proposed reward function modeling method, two other models are selected as comparison baselines, i.e., IDM and MOBIL for longitudinal and lateral behaviors respectively, and the constant velocity model. For IDM, the tuned parameters are: maximum acceleration $a_{max} = 1.3 \ m/s^2$, desired time gap $\tau = 1.2 \ s$, comfortable braking deceleration $b = 0.7 \ m/s^2$, and minimum distance $s_0 = 1.5 \ m$; and the parameters for MOBIL are: safe deceleration limitation $b_{safe}=2 \ m/s^2$, politeness factor $p=0.01$, and lane-changing decision threshold $a_{th}=0.2 \ m/s^2$. They are applied in the same testing conditions as the reward function modeling method, and only one trajectory is generated as they are deterministic methods. The metric to quantify the modeling accuracy is the average human likeness on trajectories. The results of the modeling accuracy of different models are shown in Fig. \ref{fig7}(b), in which the boxplot display the summary of the 100 different vehicles in the testing conditions.

The results in Fig. \ref{fig7}(a) indicate that the learned reward functions show acceptable robustness in the testing conditions. There is only a slight deterioration in human likeness in the testing conditions, which means the learned reward function is robust in selecting the candidate trajectories close to human driving ones in the untrained conditions. The personalized reward modeling method outperforms the general modeling method in terms of robustness, as the general modeling method shows worse mean accuracy and higher variance in the testing conditions. Fig. \ref{fig7}(b) reveals that personalized modeling is more close to human driving behavior and thus demonstrates smaller errors to the ground-truth trajectories ($Mean=2.066 \ m$). A notable reduction in human likeness is found when turning to general modeling ($Mean=2.681 \ m$), but it is still significantly better than the IDM+MOBIL model ($Mean=4.504 \ m$) and the constant velocity model ($Mean=4.986 \ m$). Additionally, the difference in human likeness between the personalized modeling and general modeling is statistically significant found by the t-test ($p<0.001$). These findings suggest that a general reward function can encode the basic requirements and common preference of human driving behaviors, whereas the personalized reward function is able to express the diverse human driving preferences and thereby achieves better performance in fitting personalized driving behavior.


\subsection{Effects of interaction factors}
We have included the interaction factors among human drivers in this paper, reflected in either taking into account the speed loss on other vehicles caused by the ego vehicle's movement or the modeling of behaviors of surrounding vehicles. We now investigate how these two interaction factors could affect the modeling accuracy of human driving behaviors in terms of training performance and generalization capability. The former aspect is achieved by removing the interaction feature (Eq. (\ref{eq16})) from the reward function, while the latter one is to let the surrounding vehicles stick to their original trajectories instead of being overridden by reactive behaviors. The same 100 target vehicles in the previous subsection are selected and the results are shown in Table \ref{tab1}. The metrics, human likeness and training log-likelihood, are averaged first by the trajectories of a vehicle and then by different vehicles.

\begin{table}[htp]
\caption{Comparison of training and testing performance for personalized modeling with regard to interaction factors}
\label{tab1}
\centering
\resizebox{\linewidth}{!}{%
\begin{tabular}{@{}ccccc@{}}
\toprule
Method          & \begin{tabular}[c]{@{}c@{}} Human likeness \\ (training) [m]\end{tabular} & \begin{tabular}[c]{@{}c@{}} Log-likelihood  \\ (training) \end{tabular} & \begin{tabular}[c]{@{}c@{}} Human likeness \\ (testing) [m]\end{tabular} \\ \midrule
Proposed                    & 1.572         & -2.062           &\textbf{2.066}  \\
W/o interaction awareness   & 1.708         & -2.145           & 2.199          \\
W/o reactive response       &\textbf{1.533} & \textbf{-2.016}  & 2.145          \\
\bottomrule
\end{tabular}
}
\end{table}

It is apparent from Table \ref{tab1} that removing the interaction awareness in the reward function would impair the modeling accuracy, which suggests that the interaction or courtesy factor is of importance in modeling naturalistic human driving behaviors. Another finding is that although not simulating the reactive behaviors of surrounding vehicles may produce better training performance in terms of both human likeness and likelihood, its generalization ability is compromised and testing performance is worse than the proposed method. The issue is possibly caused by the biased estimation of the partition function. For the sampling-based IRL method, generating accurate and possible samples that agree with the realistic human behaviors is key to approximate the partition function. In the environment model, if all other vehicles go along their fixed paths, it will make some of the sampled trajectories less likely since these trajectories could cause a crash or become too risky. However, those trajectories are still possible in the real world setting because other drivers can adapt to the ego vehicle's actions and therefore the stochasticity of human driving behavior is ignored. This could remove or underestimate some sampled trajectories when approximating the partition function, and thus bias the estimation to fit the training data and consequently leading to compromised generalization ability. Therefore, it is reasonable to simulate other vehicles' responses to the sampled trajectories in order to approximate the partition function and learn the parameters of the cost function more accurately.

For the general modeling assumption, a total of 150 trajectories from 20 vehicles are selected as training data to learn a general reward function and then the learned cost function is utilized to select the candidate trajectories in the same testing conditions as the personalized modeling method. The human likeness and log-likelihood of the training process are displayed in Table \ref{tab2}. Note that the training log-likelihood is averaged over all training trajectories.

\begin{table}[htp]
\caption{Comparison of training and testing performance for general modeling with regard to interaction factor}
\label{tab2}
\centering
\resizebox{\linewidth}{!}{%
\begin{tabular}{@{}ccccc@{}}
\toprule
Method          & \begin{tabular}[c]{@{}c@{}} Human likeness \\ (training) [m]\end{tabular} & \begin{tabular}[c]{@{}c@{}} Log-likelihood  \\ (training) \end{tabular} & \begin{tabular}[c]{@{}c@{}} Human likeness \\ (testing) [m]\end{tabular}  \\ \midrule
Proposed                  & 2.231          & -2.411          & \textbf{2.681}  \\
W/o interaction awareness & 2.473          & -2.474          & 3.410           \\
W/o reactive response     & \textbf{2.161} & \textbf{-2.381} & 3.174           \\
\bottomrule
\end{tabular}
}
\end{table}

The findings in Table \ref{tab2} are consistent with those in Table \ref{tab1}, suggesting that ignoring interaction awareness would lower the modeling accuracy and likelihood of human demonstrations both in training and testing, and not simulating the responses of other vehicles could produce better training performance but undermine the generalization capability.

Moreover, we now consider another setting where the ground truth of the future behavior of the surrounding vehicles is not available and use IDM and MOBIL models to forecast the trajectories of the nearby vehicles without the reliance on the log-replay data. First, we use the same training set to learn the reward function with the forecasting model instead of log-replay data to investigate the effect of the environment model. Second, using the same testing set, we test the previously learned reward function for trajectory planning along with the forecasting model in an open-loop way, i.e., comparing the human-likeness of the planned (generated) trajectories and ground truth human driving trajectories. This reflects how the learned reward functions can be used in planning and decision-making processes in real-world application scenarios. For calculating the interaction feature, the influenced surrounding vehicles will be the ones following behind the ego vehicle in the same lane, which can also react sequentially to the action of the ego vehicle. The results are given in Table \ref{tab3}, which reveal that the human-likeness of applying the personalized reward function degrades in this scenario as a result of inaccurate forecasting but still outperforms that of using the general reward function. It is primarily due to the inaccurate forecasting of the surrounding vehicles and thus the inaccurate estimation of the risk features, which often have higher weights for most individuals. However, the general reward function is shown to be more robust against the change of the environment model as the human-likeness remains at nearly the same level. It indicates that the accuracy of using the personalized reward function in the planning module is sensitive to the accuracy of the forecasting model while the generalized reward function is more robust to the accuracy of the forecasting model. Besides, using only the forecasting model instead of log-replay data to learn the reward function would bring a decline in accuracy in training but could still deliver a similar performance in testing. 

\begin{table}[htp]
\caption{Evaluation of the learned reward function with the forecasting model}
\label{tab3}
\centering
\resizebox{\linewidth}{!}{%
\begin{tabular}{@{}ccccc@{}}
\toprule
Method          & \begin{tabular}[c]{@{}c@{}} Human likeness \\ (training) [m]\end{tabular} & \begin{tabular}[c]{@{}c@{}} Log-likelihood  \\ (training) \end{tabular} & \begin{tabular}[c]{@{}c@{}} Human likeness \\ (testing) [m]\end{tabular}  \\ \midrule
Proposed (personalized)             & --             & --               & \textbf{2.316} \\
Proposed (general)                  & --             & --               & 2.722          \\
W/ forecasting model (personalized) & 1.938          & -2.181           & 2.354          \\
W/ forecasting model (general)      & 2.266          & -2.429           & 3.158          \\
\bottomrule
\end{tabular}
}
\end{table}

\subsection{Discussions}
The application of the proposed driving behavior modeling method is primarily on the planning and decision-making module of an AV for personalized driving experiences. It is very promising to learn a personalized cost function from naturalistic human driving data offline and integrate the learned cost function into the trajectory planning module, eventually achieving personalized driving experience. Another application is to predict the motion of surrounding vehicles. The reward functions of other vehicles can be inferred online through an offline dataset containing a distribution of cost functions for different driving styles \cite{sun2020expressing} or even acquired via vehicle-to-vehicle communications. Due to the mutual influence of agents, the trajectories of all interacting vehicles should be predicted. This would significantly increase the computation time, but the prediction process can be accelerated by reducing the sampling space or the number of target vehicles and parallelizing the sampling process.

It is worth noting that the recovered reward function in this paper may not be the same as the reward function in classic reinforcement learning, which directly drives the behavior of an agent interacting with the environment. Since we leverage the assumption on human driving behaviors, the reward function is only used to score the generated candidate trajectories, and thereby the learned reward function is somehow tailored to fit our problem setting. Whether the recovered reward function is usable in the classical sense of reinforcement learning requires further investigation.

Moreover, several limitations need to be acknowledged. First of all, the assumption of a linear reward function with time-invariant weights may not hold in real-world scenarios and the hand-crafted features cannot fully represent the factors involved in human driving behaviors. Secondly, the trajectory sampling space in this paper may not be enough for covering all possible maneuvers. This can be improved to include more complicated driving behaviors by increasing the targets in the sampling space and diversifying the planning horizon. For example, we can segment the 5-second planning horizon into five 1-second intervals and sampling a target state for each interval and more target states can be added to generate complex actions. Therefore, future work may focus on using a neural network to parameterize the reward function that maps from raw sensory states to reward value, which could help deal with nonlinear reward function modeling and improve the expression ability. Another focus is to refine the trajectory sampling method to capture more diverse behavior driving behaviors and achieve a more accurate estimation of the partition function.

\section{Conclusions}
\label{sec5}
In this study, we utilize the internal reward function-based approach to model driving behavior from naturalistic human driving data in the highway driving scenario. To enable the maximum entropy IRL algorithm to be used to infer the reward function in our problem, we propose a structural assumption about human driving behaviors that focuses on discrete latent intentions, which govern the continuous low-level control actions. According to our assumption, we first use a polynomial trajectory sampler to generate candidate trajectories covering the high-level decisions (lane-changing and lane-keeping) and the desired speed, while the generated trajectories are used to approximate the partition function in the maximum entropy IRL framework. An environment model is built up to predict the trajectories of the surrounding vehicles and evaluate the outcomes of the generated trajectories, and thus the speed loss of surrounding vehicles due to the change-of-course behavior of the ego agent can be incorporated into the reward function. We apply the proposed method to learn the personalized reward functions of different human drivers in the NGSIM dataset and interpret their driving decisions with the learned reward functions qualitatively. The quantitative results on 100 vehicles show that the personalized modeling method outperforms the general modeling method in terms of both robustness and human likeness. Moreover, the reward-function-based models significantly outperform the IDM+MOBIL model and constant velocity model. We also find out that without simulating the response actions of the vehicles influenced by the ego vehicle's generated trajectory could produce better training results but compromise the generalization ability, and without interaction awareness (the ego vehicle's action imposing speed loss on other vehicles) could also lower the modeling accuracy. Moreover, we investigate applying personalized reward functions with a forecasting model in the trajectory planning process and find out that the accuracy of personalized planning relies on the accuracy of the forecasting model but still outperforms that with a general reward function.

\ifCLASSOPTIONcaptionsoff
  \newpage
\fi

\bibliographystyle{IEEEtran}
\bibliography{reference}

\begin{IEEEbiography}[{\includegraphics[width=1in,height=1.25in,clip,keepaspectratio]{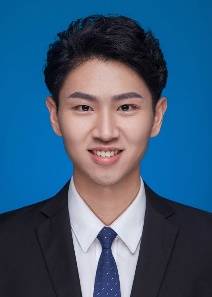}}]{Zhiyu Huang} received his B.E. degree from the School of Automobile Engineering, Chongqing University, Chongqing, China, in 2019. He is currently pursuing his Ph.D. degree with the School of Mechanical and Aerospace Engineering, Nanyang Technological University, Singapore. His current research interests include machine learning for prediction and decision-making in automated driving and human-machine interactions.
\end{IEEEbiography}

\begin{IEEEbiography}[{\includegraphics[width=1in,height=1.25in,clip,keepaspectratio]{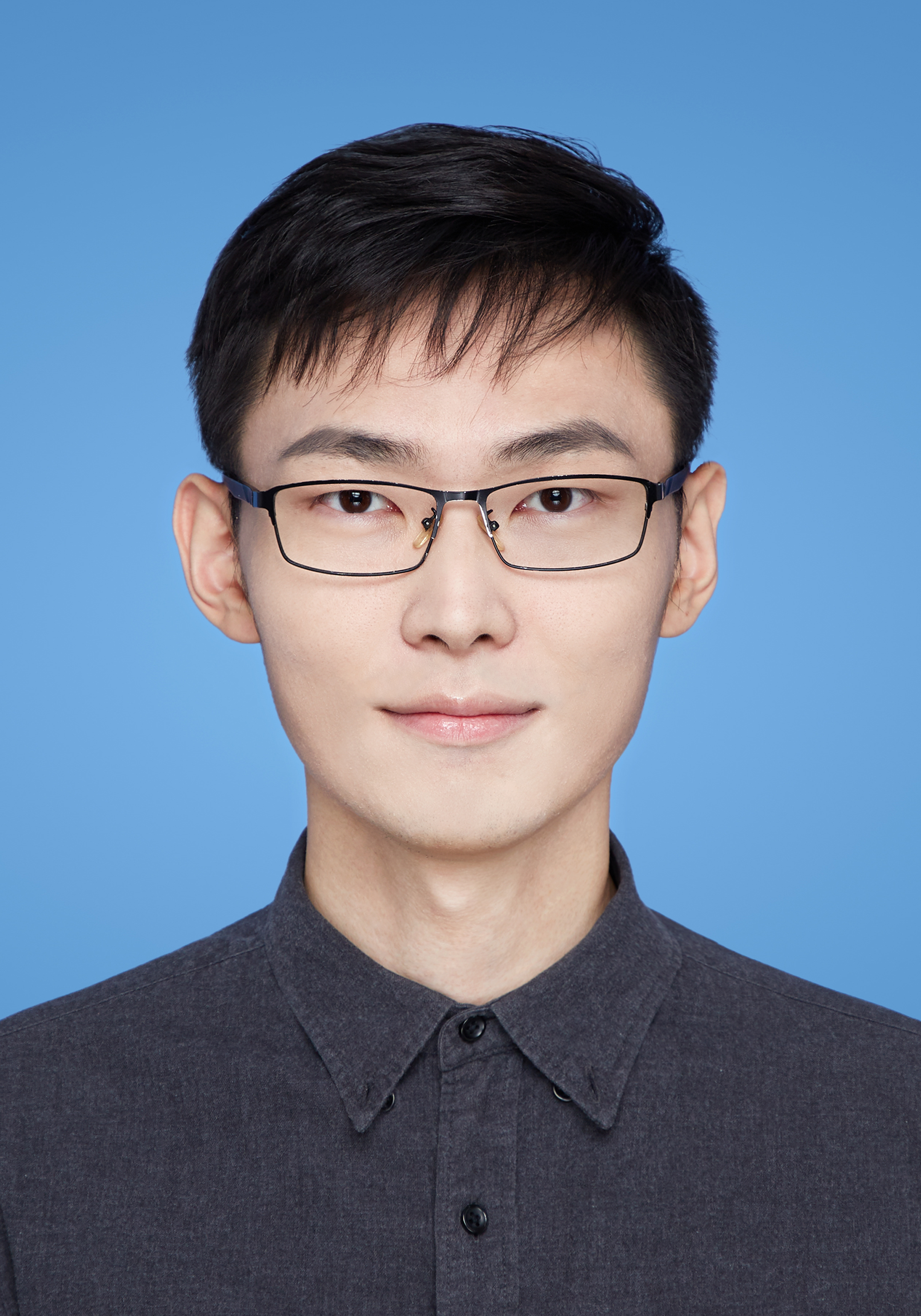}}]{Jingda Wu} received his B.S. (2016) and M.S. (2019) in mechanical engineering from Beijing Institute of Technology, China. He is currently working on his Ph.D. in mechanical engineering at Nanyang Technological Univ., Singapore. His research mainly focuses on control and optimization of human machine collaborated driving, machine learning techniques, design of autonomous driving strategy, energy management of electric vehicle and Li-ion battery.
\end{IEEEbiography}

\begin{IEEEbiography}[{\includegraphics[width=1in,height=1.25in,clip,keepaspectratio]{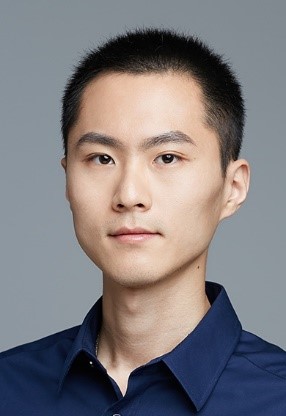}}]{Chen Lv} (M’16-SM’20) is currently an Assistant Professor at Nanyang Technology University, Singapore. He received the Ph.D. degree at the Department of Automotive Engineering, Tsinghua University, China in 2016. He was a joint Ph.D. researcher at EECS Dept., University of California, Berkeley, USA, during 2014-2015, and a Research Fellow with the Advanced Vehicle Engineering Center, Cranfield, University, U.K., during 2016 and 2018. His research focuses on advanced vehicles and human-machine systems, where he has contributed over 100 papers and obtained 12 granted patents. Dr. Lv serves as a Guest Editor for IEEE/ASME TMECH, IEEE ITS Magazine, Applied Energy, etc., and an Associate Editor/Editorial Board Member for International Journal of Vehicle Autonomous Systems, Frontiers in Mechanical Engineering, Vehicles, etc. He received the Highly Commended Paper Award of IMechE UK in 2012, the NSK Outstanding Mechanical Engineering Paper Award in 2014, the CSAE Outstanding Paper Award in 2015, Tsinghua University Outstanding Doctoral Thesis Award in 2016, the CSAE Outstanding Doctoral Thesis Award, and IEEE IV Best Workshop/Special Session Paper Award in 2018.
\end{IEEEbiography}

\end{document}